\pdfoutput=1
\documentclass[11pt]{article}

\usepackage[]{emnlp2021}

\usepackage{times}
\usepackage{latexsym}

\usepackage[T1]{fontenc}
\usepackage[utf8]{inputenc}

\usepackage{microtype}

\usepackage{amsmath}
\usepackage{diagbox}
\usepackage{booktabs}
\newcommand{\ra}[1]{\renewcommand{\arraystretch}{#1}}
\usepackage{tcolorbox}
\title{Language Models are Few-Shot Butlers}

\author{Vincent Micheli \\
  University of Geneva \\
  \texttt{vincent.micheli@unige.ch} \\\And
  François Fleuret \\
  University of Geneva \\
  \texttt{francois.fleuret@unige.ch} \\}

\begin{document}
\maketitle
\begin{abstract}
Pretrained language models demonstrate strong performance in most NLP tasks when fine-tuned on small task-specific datasets. Hence, these autoregressive models constitute ideal agents to operate in text-based environments where language understanding and generative capabilities are essential. Nonetheless, collecting expert demonstrations in such environments is a time-consuming endeavour.
We introduce a two-stage procedure to learn from a small set of demonstrations and further improve by interacting with an environment. We show that language models fine-tuned with only 1.2\% of the expert demonstrations and a simple reinforcement learning algorithm achieve a 51\% absolute improvement in success rate over existing methods in the ALFWorld environment.
\end{abstract}

\section{Introduction}
\label{sec:introduction}

Over the past few years, successive generations of language models \citep{gpt1, gpt2, gpt3} have reshaped the way we approach Natural Language Processing problems. These Transformer-based \citep{attention} networks scale to ever-increasing amounts of parameters, data and compute \citep{scalinglaws} while demonstrating impressive transfer-learning capabilities across a wide variety of benchmarks \citep{wang-etal-2018-glue, superglue, rajpurkar-etal-2018-know, reddy-etal-2019-coqa}.

However, the application of modern language models to action generation in text-based environments \citep{textworld, jericho, alfworld} remains largely unexplored \citep{yao-etal-2020-keep}. Intuitively, large-scale models leveraging strong linguistic priors should thrive in settings where observations and actions are textual \citep{survey-rl-nlp}, but several issues arise. Indeed, agents must conform to an environment's generative grammar and gameplay specificities making off the shelf transfer fail in most games. A way to overcome this problem is to acquire expert demonstrations and resort to the widely used paradigm of fine-tuning on task-specific data \citep{howard-ruder-2018-universal, gpt1, devlin-etal-2019-bert}. Nevertheless, collecting demonstrations in text-based environments requires far more time and expert knowledge than for most NLP tasks. A single demonstration includes tens of actions taken over a long time horizon to solve multiple sub-goals.

\begin{figure}[!t]
    \begin{tcolorbox}[colback=white, left=1mm, right=1mm, top=1mm, bottom=1mm]
\emph{Goal:} Rinse the egg to put it in the microwave. \\
\emph{Obs:} Looking quickly around you, you see a cabinet, a garbagecan, a coffeemachine, [...],  a stoveburner, a sinkbasin and a microwave. \\
\emph{Action:} go to sinkbasin \\
\emph{Obs:} You arrive at sinkbasin. You see a butterknife, a potato, a spoon and a tomato. \\
\emph{Action:} go to garbagecan \\
\emph{Obs:} You arrive at garbagecan. You see an egg. \\
\emph{Action:} take egg from garbagecan \\
\emph{Obs:} You pick up the egg from the garbagecan. \\
\emph{Action:} go to sinkbasin \\
\emph{Obs:} You arrive at sinkbasin. You see a butterknife, a potato, a spoon and a tomato. \\
\emph{Action:} clean egg with sinkbasin \\
\emph{Obs:} You clean the egg using the sinkbasin. \\
\emph{Action:} go to microwave \\
\emph{Obs:} You arrive at microwave. The microwave is closed. \\
\emph{Action:} open microwave \\
\emph{Obs:} You open the microwave. \\
\emph{Action:} put egg in/on microwave
    \end{tcolorbox}
    \caption{Example of a human-annotated, out-of-distribution task instance solved by $\text{GPT}^{\star}\textsubscript{partial}$.}
    \label{fig:example}
\end{figure}

In this work, we  propose a two-stage procedure to address these issues and develop language models acting as agents in text-based environments. First, we train language models to imitate a few dozens of expert demonstrations in order to respect an environment's grammar and acquire basic gamesense. Second, we let the models interact with the environment and iteratively treat successful trajectories as additional expert demonstrations for further fine-tuning. We demonstrate the effectiveness of our approach in the recently introduced ALFWorld environment \citep{alfworld}\footnote{ALFWorld aligns both text and embodied environments, but here we only refer to the text environment.}, which was designed with an extensive set of tasks and expert demonstrations.

In summary, our contributions are the following:
\begin{enumerate}
    \item We show that language models fine-tuned on thousands of expert demonstrations considerably outperform current methods in the ALFWorld environment.
    \item We achieve strong results with a fraction of the demonstrations by combining imitation and reinforcement learning algorithms.
    \item We illustrate the robustness of the models developed to human-annotated goals in realistic scenarios.
\end{enumerate}

\section{Methods}

\subsection{Background: goal-based textual environments}

A goal-based textual environment can be represented as a partially observable Markov decision process $P=(\mathcal{S}, \mathcal{O}, \mathcal{A}, \mathcal{G}, R, T, M)$ where observations, actions and goals are specified in natural language. In state $s_t \in \mathcal{S}$, an agent takes action $a_t \in \mathcal{A}$ conditioned on context $c_t = (g, o_0, a_0, ..., o_t)$. It receives reward $r_t = R(s_t, a_t, g)$, which is an indicator variable for the completion of goal $g \in \mathcal{G}$, and a new observation $o_{t+1} = M(T(s_t, a_t))$, where $M\colon \mathcal{S} \rightarrow \mathcal{O}$ is a mapping from states to observations and $T\colon \mathcal{S} \times \mathcal{A} \rightarrow \mathcal{S}$ is the transition function.

\subsection{Learning from demonstrations}

A demonstration $d$ consists of a sequence of observations and actions $(o_0, a_0, o_1, a_1, ..., o_T, a_T)$ for reaching goal $g$ based on contexts $(c_0, c_1, ..., c_T)$. We consider a dataset $\mathcal{D}$ of $N$ demonstrations. A parameterized model $p_\theta$ is trained to minimize the mean demonstration loss $\mathcal{L}_\mathcal{D} = \frac{1}{N} \sum_{i=1}^{N} \mathcal{L}_{d_i}$ with
\begin{equation}
     \mathcal{L}_{d_i} = - \sum_{t=0}^{T} \log p_{\theta}(a_t | c_t).
\end{equation}
As noted by \citet{yao-etal-2020-keep}, 
\begin{equation}
\log p_{\theta}(a | c) = \sum_{j=1}^{m} \log p_{\theta}(a^{j} | a^{<j}, c),
\end{equation}
where $a^{j}$ is the j-th token generated in action $a$ of length $m$.

We use a per-demonstration loss instead of a per-action loss to reduce computational costs. Indeed, with this formulation a Transformer-based autoregressive model can leverage previous computations when considering a new context from the same demonstration. In addition, early experiments suggested that a per-demonstration loss does not harm performance.

We call \emph{action modeling} the process of minimizing the mean demonstration loss, which is conceptually very similar to language modeling, except that we only maximize the likelihood of action tokens instead of maximizing the likelihood of the full trajectory $c_T$.

\begin{table*}\centering
\ra{1.2}
\begin{tabular}{@{}lcccccc@{}}\toprule
& \multicolumn{2}{c}{ALFWorld goals} & \phantom{abc}& \multicolumn{2}{c}{Human goals}\\
\cmidrule{2-3} \cmidrule{5-6}
& Seen split & Unseen split && Seen split & Unseen split\\ \midrule
Seq2Seq \citep{alfworld} & 10 & 9 &&  & \\
BUTLER \citep{alfworld} & 40 & 37 &&  & \\
GPT\textsubscript{partial} & 47 & 40 && 17 & 22\\
$\text{GPT}^{\star}\textsubscript{partial}$ & 69 & 60 && 32 & 37\\
GPT & 91 & 95 && 42 & 57\\
\bottomrule
\end{tabular}
\caption{Success percentages per evaluation split (in-distribution and out-of-distribution) with and without human-annotated goals. GPT\textsubscript{partial} and GPT are GPT2-based models fine-tuned with action modeling on 42 and 3553 demonstrations, respectively. $\text{GPT}^{\star}\textsubscript{partial}$ corresponds to the former model subsequently trained with iterated action modeling in ALFWorld. Our results are averaged over 5 seeds. Standard deviations are upper bounded by 9 for GPT\textsubscript{partial}, 8 for $\text{GPT}^{\star}\textsubscript{partial}$, and 3 for GPT.}
\label{tab:results}
\end{table*}

\subsection{Learning from interactions}

While action modeling is a powerful training objective, a model learning from demonstrations is ultimately limited by the size of the training set. To circumvent this issue, we propose an \emph{iterated action modeling} (IAM) algorithm:
\begin{enumerate}
    \item A language model pretrained on expert demonstrations is tasked to solve a batch of goals in the environment.
    \item The language model is further fine-tuned with action modeling on successful trajectories.
\end{enumerate}
The key advantage of this algorithm is that we can easily combine imitation learning with reinforcement learning since we are optimizing the same objective over two distinct sources of data: demonstrations and successful attempts. Moreover, the extensively pretrained language/action modeling head is kept during reinforcement learning instead of initializing a new RL-specific head from scratch, which was shown to lead to better performance in NLP tasks \citep{lm-head}.

\section{Experiments}

Experiments were implemented with the Transformers \citep{wolf-etal-2020-transformers} and PyTorch \citep{pytorch} libraries and were conducted on an NVidia RTX 3090.\footnote{Source code and links to models available at: \url{https://github.com/vmicheli/lm-butlers}}

\subsection{Environment and dataset}

ALFWorld \citep{alfworld} is a goal-based textual environment mirroring the embodied ALFRED benchmark \citep{alfred} with the TextWorld game engine \citep{textworld}. The environment was created with the aim of learning high-level language policies inside of it and transferring them to the embodied setting. ALFWorld inludes 6 tasks that are compositional and require multiple sub-goals to be solved over various time horizons. Any string of words constitute a valid action making the action space unbounded and the training of a policy consequently difficult. In total there are 3553 training task instances \{task-type, object, receptacle, room\}, 140 in-distribution evaluation task instances (seen split) and 134 out-of-distribution evaluation task instances (unseen split). A task instance specifies the type of the task to solve, the object to interact with, the receptacle where the object should be put and the room layout (e.g. \{heat and place, egg, countertop, kitchen 12\}). Besides, each training task instance in ALFWorld comes with an expert demonstration, enabling the development of imitation learning agents.

\subsection{Training}

We train two GPT2-medium (345M parameters) \citep{gpt2} models with action modeling on the set of demonstrations. The first model, GPT, has access to the full set of demonstrations while the second model, GPT\textsubscript{partial}, only has access to 42 demonstrations.
GPT\textsubscript{partial} is subsequently trained with iterated action modeling in the environment and is then denoted as $\text{GPT}^{\star}\textsubscript{partial}$. When interacting with the environment, models greedily decode actions token-per-token until an end of action token is reached. See Appendix \ref{sec:training-details} for training details.

\subsection{Evaluation}
\label{sec:evaluation}

We select model checkpoints according to their evaluation performance on the seen split and further evaluate them on the unseen split. During evaluation, we employ greedy action decoding and a sliding context window which depends on the maximum number of tokens the language models can handle. This implies that the contexts given to the models consist of the goal, the first observation and as many of the previous observations and actions as possible. We compare our models with the ones developed by \citet{alfworld}:
\begin{itemize}
    \item BUTLER: trained with Dagger \citep{dagger} for 50k episodes and handling failed actions with beam search.
    %\item BUTLER\textsubscript{g}: trained with Dagger for 100k episodes.
    \item Seq2Seq: trained with the full set of demonstrations.
\end{itemize}
Contrary to our approach, these models do not encapsulate prior linguistic knowledge except from pretrained word embeddings.

\subsection{Robustness}

In ALFWorld, goals follow a generative grammar specific to the environment, e.g.\ "put a hot apple in fridge". However, when interacting with autonomous agents, humans may formulate goals that deviate from this grammar, e.g.\ "warm up apple to put in fridge". The ability to generalize to human-annotated goals is quantitatively assessed with crowd-sourced goal annotations \citep{alfred, alfworld}. We evaluate the best performing models from Section \ref{sec:evaluation} on the human-annotated seen and unseen splits.

\section{Results}

\subsection{Language models strongly outperform existing methods in ALFWorld}

We report the entirety of the results in Table \ref{tab:results}. GPT achieves success rates of 91\% and 95\%, respectively, on the seen and unseen splits. That is, absolute improvements of 81\% and 86\% over the Seq2Seq model trained on the same data. Even when compared to BUTLER, trained with 14 times more expert-guided demonstrations and manually handling failed actions, we observe absolute improvements of 51\% and 58\%. GPT\textsubscript{partial} is also competitive with BUTLER and outperforms the Seq2Seq model with only 0.07\% and 1.2\% of the expert demonstrations available. However, there remains a large performance gap between the two GPT2-based models.

\subsection{Iterated action modeling retains most of the performance with few demonstrations}

With iterated action modeling, GPT\textsubscript{partial}'s performance improves by 22\% and 20\%, respectively, on the seen and unseen splits. In other words, $\text{GPT}^{\star}\textsubscript{partial}$ retains 76\% and 63\% of GPT's results with only 1.2\% of the expert demonstrations available.

\subsection{Agents with linguistic priors are robust to human-annotated goals}

Evaluation on the seen and unseen splits with human-annotated goals reveals that language models fine-tuned with action modeling on expert demonstrations and successful trajectories are capable of solving a large proportion of goals formulated in open-ended natural language. 
For example, GPT and $\text{GPT}^{\star}\textsubscript{partial}$ solve respectively 57\% and 37\% of human-annotated, out-of-distribution task instances. Figure \ref{fig:example} illustrates $\text{GPT}^{\star}\textsubscript{partial}$ solving one of these tasks.

\section{Related work}

\citet{yao-etal-2020-keep} used language models to prune the action space in text-based games. The authors introduced the ClubFloyd dataset, which contains gameplay transcripts collected over a multitude of games, and fine-tuned a GPT2-small (117M parameters) \citep{gpt2} model on that dataset for action generation. This contextual action language model (CALM) was then queried to generate a small list of action candidates based on the last few observations and actions. CALM was combined with game-specific models trained with reinforcement learning \citep{he-etal-2016-deep} to pick the best action candidate among CALM's generations. This approach aims to transfer a general-purpose language model across multiple new environments without game-specific imitation or reinforcement learning. In our work, we optimize for performance instead of generalization by training language models with game-specific demonstrations and interactions. In fact, preliminary experiments with CALM in ALFWorld reveal that the model is unable to produce valid actions both in terms of grammar and task completion.

Goal-conditioned supervised learning \citep{gcsl} treats every trajectory as an expert demonstration for reaching the final state encountered in that same trajectory. This hindsight goal-relabeling is possible because there exists a straightforward mapping between goals and states in the environments considered (i.e.\ the identity map). In ALFWorld, learning such a mapping is highly non-trivial and constitutes another research direction for extending existing methods \citep{higher} to this environment. Therefore, during iterated action modeling we only consider successful trajectories as expert demonstrations and initialize the agent with a few demonstrations in order to start the RL procedure with a non-zero success rate.

\section{Conclusion}

We developed new agents for text-based environments with pretrained language models. These agents acquired game knowledge through demonstrations and interactions to drastically outperform current methods in the ALFWorld environment. While we investigated learning under the standard fine-tuning paradigm, more sophisticated approaches could be explored \citep{seq-pet} and recent works \citep{gpt3, prompt-calibrate} even suggest that scaled-up and carefully calibrated models achieve great downstream results without requiring any parameter updates. Thus, in the near future one can imagine language models solving text-based environments with only a few demonstrations for priming.

\section*{Acknowledgments}

We thank the ALFWorld team for their technical support regarding the environment.

Vincent Micheli was supported by the Swiss National Science Foundation under grant number FNS-187494 "Meaningful Human Control of Security Systems -- Aligning International Humanitarian Law with Human Psychology".

\bibliographystyle{acl_natbib}
\bibliography{anthology,custom}

\newpage

\appendix

\section{Hyperparameters and sample selection}
\label{sec:training-details}

We do not leverage any (potentially large) held-out set of demonstrations to tune hyperparameters or learning objectives. As mentioned in \ref{sec:evaluation}, we solely optimize the success rate over a small set of validation task instances that we can freely query rather than a validation loss on held-out examples. Hyperparameters for the action modeling and iterated action modeling experiments are displayed in Table \ref{tab:am-params} and Table \ref{tab:iam-params}. For the action modeling experiments with GPT\textsubscript{partial}, we randomly select 7 demonstrations per task-type from the pool of 3553 demonstrations.

\begin{table}[h!]
\centering
\begin{tabular}{lcc}
\hline \textbf{Hyperparameter} & \textbf{GPT} & \textbf{GPT\textsubscript{partial}} \\ \hline
Epochs & \{\textbf{10}, 20\} & 100 \\
Batch size & 1 & 1 \\
Gradient acc. steps & 8 & 7 \\
Learning rate & 5e-5 & \{1e-5, \textbf{5e-5}\}\\
LR schedule & Linear & Constant \\
Adam $\beta_1$ & 0.9 & 0.9 \\
Adam $\beta_2$ & 0.999 & 0.999 \\
Max gradient norm & 1.0 & 1.0 \\
Dropout & 0.1 & 0.1 \\
Max sequence length & 1000 & 1000 \\
\hline
\end{tabular}
\caption{Action modeling hyperparameters.}
\label{tab:am-params}
\end{table}

\begin{table}[h!]
\centering
\begin{tabular}{lc}
\hline \textbf{Hyperparameter} & \textbf{$\text{GPT}^{\star}\textsubscript{partial}$} \\ \hline
Iterations & 20 \\
Episodes per iteration & \{100, 200, \textbf{400}\} \\
Batch size & 1 \\
Gradient acc. steps & 8 \\
Learning rate & \{1e-6, \textbf{1e-5}, 5e-5\} \\
LR schedule & Constant \\
Adam $\beta_1$ & 0.9\\
Adam $\beta_2$ & 0.999 \\
Max gradient norm & 1.0 \\
Dropout & 0.1 \\
Max action length & 20 \\
Max sequence length & 1000 \\
Action selection & Sampling \\
\hline
\end{tabular}
\caption{Iterated action modeling hyperparameters.}
\label{tab:iam-params}
\end{table}

\section{Performance as a function of the number of training demonstrations}

\begin{figure}[h!]
  \centering
  \includegraphics[width=0.48\textwidth]{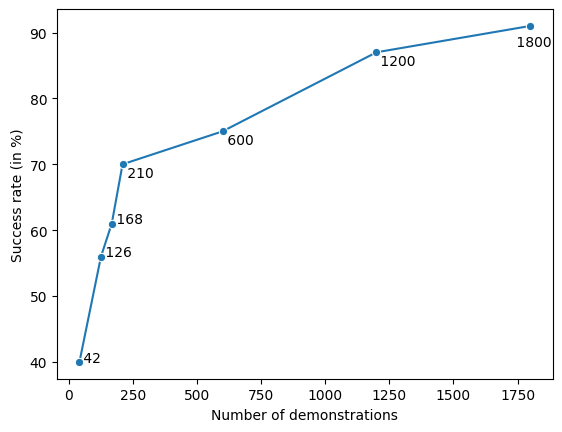}
  \caption{Unseen split performance as a function of the number of training demonstrations.}
  \label{fig:learning-curve}
\end{figure}

In Figure \ref{fig:learning-curve}, we provide a curve of model performance as a function of the number of training demonstrations for the action modeling stage.

Around 168 demonstrations are necessary to achieve a success rate equivalent to that of $\text{GPT}^{\star}\textsubscript{partial}$. In other words, adding the iterated action modeling procedure brings improvements similar to those we would get if we multiplied the number of demonstrations by 4.

\section{Input representation}

In practice, a context is formed in the following way:

\begin{enumerate}
    \item Append the goal to the first observation.
    \item Preprend modality strings "[STATE]" and "[ACTION]" to observations and actions.
    \item Concatenate past observations and actions in a single string of text.
\end{enumerate}

\end{document}